%% file: humanref.tex
\documentclass[10pt,twocolumn,letterpaper]{article}

\usepackage[pagenumbers]{cvpr} % To force page numbers, e.g. for an arXiv version

\usepackage{makecell}
\usepackage{cuted}
\newcommand*{\dif}{\mathop{}\!\mathrm{d}}

\newcommand{\ignore}[1]{}

\title{HumanRef: Single Image to 3D Human Generation via Reference-Guided Diffusion}

\author{
Jingbo Zhang, Xiaoyu Li, Qi Zhang, Yanpei Cao, Ying Shan, and Jing Liao 
\thanks{corresponding author.} 
\thanks{J. Zhang and J. Liao are with Department of Computer Science, City University of Hong Kong. E-mail: jbzhang6-c@my.cityu.edu.hk, jingliao@cityu.edu.hk. X. Li, Q. Zhang, Y. Cao, and Y. Shan are with Tencent AI Lab.} 
}

\begin{document}
\maketitle

\begin{strip}
\setlength\tabcolsep{0.5pt}
\centering
    \begin{tabular}{c}
        \includegraphics[width=0.99\textwidth]{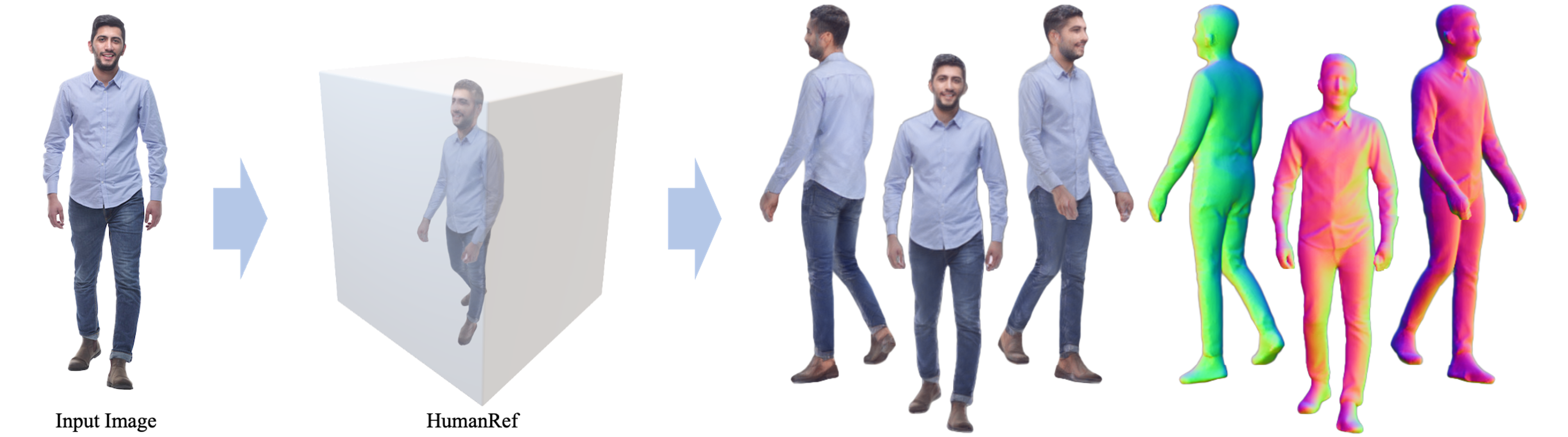}
    \end{tabular}
    % \vspace{-0.1in}
    \captionof{figure}{We propose HumanRef, a reference-guided 3D human generation framework. Our HumanRef is capable of generating 3D clothed human with realistic, view-consistent texture and geometry from a single image input.}
\label{fig:teaser}
\end{strip}

\input{sec/0_abstract}    
\input{sec/1_intro}

\input{sec/2_related}
\input{sec/3_method}
\input{sec/4_exp}

\input{sec/5_conclusion}

{
    \small
    \bibliographystyle{ieeenat_fullname}
    \bibliography{main}
}


\end{document}

%% file: sec/0_abstract.tex
\begin{abstract}
Generating a 3D human model from a single reference image is challenging because it requires inferring textures and geometries in invisible views while maintaining consistency with the reference image. Previous methods utilizing 3D generative models are limited by the availability of 3D training data. Optimization-based methods that lift text-to-image diffusion models to 3D generation often fail to preserve the texture details of the reference image, resulting in inconsistent appearances in different views. In this paper, we propose HumanRef, a 3D human generation framework from a single-view input. To ensure the generated 3D model is photorealistic and consistent with the input image, HumanRef introduces a novel method called reference-guided score distillation sampling (Ref-SDS), which effectively incorporates image guidance into the generation process. Furthermore, we introduce region-aware attention to Ref-SDS, ensuring accurate correspondence between different body regions. Experimental results demonstrate that HumanRef outperforms state-of-the-art methods in generating 3D clothed humans with fine geometry, photorealistic textures, and view-consistent appearances. We will make our code and model available upon acceptance.

%The realistic retrieval of 3D clothed humans from a single image has garnered significant attention in the fields of computer vision and graphics. However, existing methods face challenges in generating clear, realistic, and view-consistent textures in occluded regions. To address this issue, we propose HumanRef, a unified reference-guided human generation framework that achieves realistic 3D clothed human generation. To ensure the generated texture remains consistent with the input image, HumanRef introduces a novel region-aware reference-guided score distillation sampling (Ref-SDS) method, which effectively incorporates image information into the generation process. Additionally, we employ a multi-step denoising strategy based on Ref-SDS to further enhance the details of the generated texture. Experimental results demonstrate that HumanRef outperforms state-of-the-art methods in generating clear, realistic, view-consistent textures, and reasonable geometry. We will make our code and model available upon acceptance.

\end{abstract}

% All authors will benefit from reading Mermin's description of how to write mathematics:
% \url{http://www.pamitc.org/documents/mermin.pdf}.

%% file: sec/1_intro.tex
\section{Introduction}
\label{sec:intro}
% \etal.  {\em text}

Clothed human reconstruction from single or multi-view images has received significant attention in the fields of computer vision and graphics due to its potential applications in virtual reality, movie industry, and immersive games. Unlike reconstruction using videos~\cite{alldieck2018detailed, alldieck2018video, alldieck2019learning} or multi-view images~\cite{balan2007detailed, bhatnagar2019multi, wu2012full, vlasic2009dynamic}, which allows for inferring 3D information from multi-view inputs, reconstructing a 3D human model from a single view input is considerably more challenging. This task not only requires the reconstructed 3D clothed human to exhibit consistency in geometry and texture with the input view but also involves generating plausible geometry and texture that are not directly visible in the input. Therefore, compared to reconstruction, this task faces even greater challenges in human generation.

%Realistic 3D clothed human reconstruction and generation has attracted widespread attention in the fields of computer vision and graphics for its potential applications in VR, AR, virtual social media, and immersive games.  Unlike multi-view human reconstruction, there are two main cores in reconstructing a 3D clothed human from a single image. One is that the reconstructed 3D clothed human needs to be consistent in geometry and texture with the input view, and the other is to infer reasonable geometry and texture in the input invisible area. Therefore, inferring the 3D human from a single image encompasses both a reconstruction and a generation problem. In contrast to reconstruct both fantastic geometry and realistic texture from a single image, some other work like ICON~\cite{xiu2022icon} and ECON~\cite{xiu2023econ} pay attention more on the geometry reconstruction and deduce 3D clothed human with impressive detailed shape. However, how to deduce a 3D human with realistic and clear textures is still a matter of great concern.

One approach to generating 3D human models from single images is by training a 3D generative model using 3D scanned human datasets. However, methods in this category \cite{xiu2022icon,xiu2023econ} tend to be more successful in generating geometry rather than texture, as textures are too diverse to be learned from limited 3D data. For example, ICON \cite{xiu2022icon} and ECON \cite{xiu2023econ} primarily estimate geometry, while PaMIR \cite{zheng2021pamir} and PIFu \cite{saito2019pifu} tend to produce blurred textures when applied to in-the-wild data. Another approach to address the lack of 3D training data is to lift a 2D model pretrained on a large dataset to 3D. DreamFusion \cite{poole2022dreamfusion} is a pioneering work in this direction, leveraging the prior knowledge in a pretrained text-to-image diffusion model to supervise the optimization of 3D objects using a score distillation sampling (SDS) loss. Some follow-up works have extended this framework from text-to-3D generation to reference-guided 3D generation. This is achieved by extracting textual guidance from the reference image, either through textual inversion (e.g., RealFusion \cite{melas2023realfusion} and NeRDi \cite{deng2023nerdi}) or detailed text description using an image captioning model (e.g., Make-It-3D \cite{tang2023make}). TeCH \cite{huang2023tech} combines both textual inversion and image captioning with 3D human priors to apply the SDS framework to 3D human generation from a single image.

Despite its success in 3D human generation, TeCH is limited by the SDS loss for two reasons. First, SDS is guided by texts, but text prompts or embeddings extracted from images can only represent the global semantic information of the reference images, which cannot capture the lower-level features necessary to provide detailed textures. As a result, TeCH's generated textures in invisible views often exhibit inconsistency with the reference image. Secondly, given the same text prompt, the diffusion model can generate images with a large diversity, making it difficult for the SDS loss optimization to converge. To address this issue, the SDS loss employs a large class-free guidance (CFG) scale in each diffusion denoising step, aiming to enhance the text relevance and reduce the diversity of image generation. However, while this approach ensures some stability in the 3D generation process, it also leads to generated texture over-saturation and over-smoothing, which is a well-known problem associated with the SDS loss. Although some efforts have been made to alleviate this problem, such as the variational score distillation (VSD) proposed by ProlificDreamer~\cite{wang2023prolificdreamer}, these methods are primarily designed for text-to-3D generation and do not consider the reference image and human priors. Consequently, they are less optimal when applied to human generation from a single image.

\begin{figure}[t]
  \centering
   \includegraphics[width=0.99\linewidth]{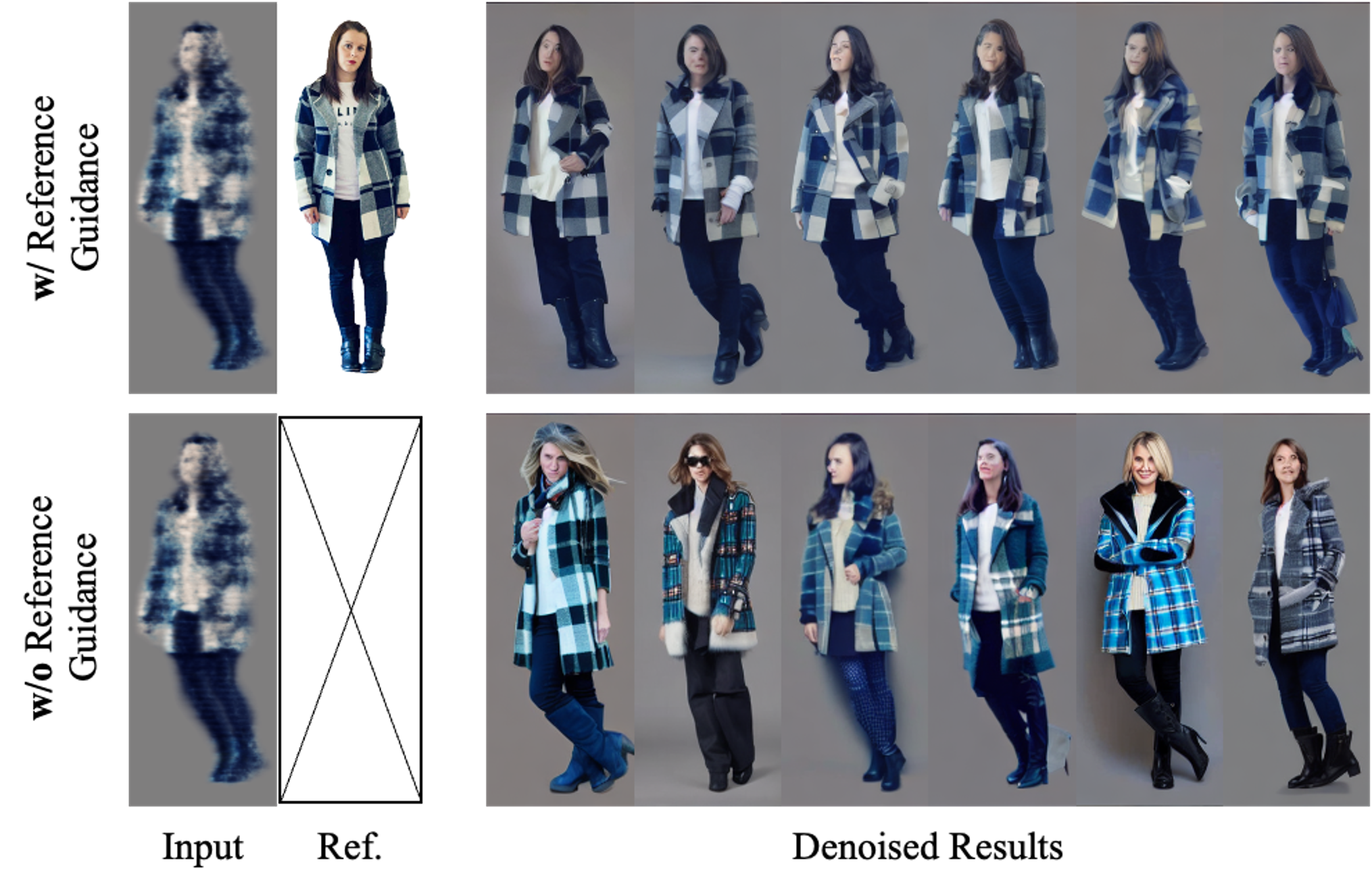}
   \caption{Showcasing the denoising process on initial input with and without reference image guidance. Starting with a coarse novel-view rendering image, we employ multiple rounds of random denoising using both reference-guided and text-guided diffusion models, highlighting the impact of reference guidance.}
   \vspace{-0.1in}
   \label{fig:img_gen}
\end{figure}

% 在本文中，我们提出了一个统一的人体生成框架，采用hash-grid的neural SDF表达，新的SDS loss， named Ref-SDS，attentio-aware多步去噪
In this paper, we present a novel framework called HumanRef for 3D clothed human generation from a reference image. Our approach utilizes a hash-encoded signed distance field (SDF) network for 3D representation and optimizes the SDF parameters from coarse to fine. We incorporate human geometry constraints and, most importantly, introduce a novel Reference-Guided Score Distillation Sampling (Ref-SDS) loss in the optimization. Unlike the vanilla SDS loss, which is guided solely by text prompts, our Ref-SDS loss fully exploits the guidance of the reference image. We inject the reference image into the diffusion model to calculate the attentions between features of the reference and generated images at each denoising step. This process guides the diffusion model in generating results that better preserve the visual appearances of the reference image, as demonstrated in Figure \ref{fig:img_gen}. Consequently, our method is superior in generating view-consistent results matching the reference image. Moreover, by distilling less diverse images from the diffusion model, our Ref-SDS loss converges more easily, enabling us to reduce the CFG scale and generate more photo-realistic textures. To further incorporate human priors, we introduce Region-Aware Attentions for Ref-SDS. We employ human parsing to segment images into body regions, enhancing the attentions of corresponding regions (e.g., head to head) and suppressing the attention of non-corresponding regions when calculating the attentions between features of the reference and generated images. This refinement improves the precision of image guidance in our Ref-SDS. Thanks to these design choices, our HumanRef framework can generate 3D clothed humans with fine geometry, photorealistic textures, and view-consistent appearances from a single reference image.

To sum up, our contributions are three-fold as below:
\begin{itemize}
\item We propose HumanRef, a coarse-to-fine optimization framework, for 3D clothed human generation from a single image. It unifies the optimization process of geometry and appearance in a single SDF representation and does not introduce additional 3D representation and optimization stages.
\item We propose a novel Reference-Guided Score Distillation Sampling (Ref-SDS) method for 3D generation. Ref-SDS injects image-level guidance into the denoising process of a pretrained diffusion model, resulting in the production of more photorealistic and view-consistent 3D results.
\item We introduce region-aware attention to Ref-SDS for 3D human generation, enhancing the precision of image guidance by ensuring accurate correspondence between different body regions.
\end{itemize}

%% file: sec/2_related.tex
\section{Related Work}
\label{sec:related}

%-------------------------------------------------------------------------
\subsection{3D Clothed Human Reconstruction}

Unlike human pose and shape estimation methods~\cite{guan2009estimating, pavlakos2018learning, bogo2016keep, omran2018neural, li2021hybrik, goel2023humans} that leverage a parametric body model for naked body reconstruction, clothed human reconstruction focuses on 3D humans with clothes, entailing more intricate details. This task has been explored in video~\cite{alldieck2018detailed, alldieck2018video, alldieck2019learning} or multi-view settings~\cite{balan2007detailed, bhatnagar2019multi, wu2012full, vlasic2009dynamic} for additional reconstruction constraints. However, the hardware requirements for additional inputs limit practical usage. Consequently, efforts have been made to recover 3D clothed humans from a single image~\cite{saito2019pifu, saito2020pifuhd, xiu2022icon, xiu2023econ, zheng2021pamir, corona2023structured, huang2020arch, alldieck2022photorealistic, liao2023high}. Notably, PIFu~\cite{saito2019pifu} digitizes detailed clothed humans by inferring 3D geometry and texture from a single image, while PIFuHD~\cite{saito2020pifuhd} introducing a coarse-to-fine framework for high-resolution geometry reconstruction. PaMIR~\cite{zheng2021pamir} combines the parametric body model with a deep implicit function, while ICON~\cite{xiu2022icon} and ECON~\cite{xiu2023econ} recover fine geometry by inferring detailed clothed human normals. PHORHUM~\cite{alldieck2022photorealistic} and S3F~\cite{corona2023structured} estimate albedo and shading information for relighting during reconstruction. Despite these advancements, such methods struggle to achieve clear and realistic textures, particularly for unseen areas in the input image. In this work, we leverage diffusion prior to synthesize high-quality, consistent textures for these invisible areas.

\subsection{Diffusion Models}
Diffusion models~\cite{sohl2015deep, ho2020denoising} are latent-variable generative models that have garnered significant attention due to their impressive generation results. It consists of a forward process that slowly removes structure from data by adding noise and a reverse process or generative model that slowly adds structure from noise. To improve the performance of diffusion models, denoising diffusion implicit models~\cite{song2020denoising} propose to use non-Markovian diffusion processes to reduce the generation steps and~\cite{dhariwal2021diffusion} proposes classifier guidance to improve the sample quality using a classifier to trade off diversity for fidelity. While~\cite{ho2022classifier} introduces classifier-free guidance by mixing the score estimates of a conditional diffusion model and a jointly trained unconditional diffusion model. Benefited from the scalability of the diffusion models and large-scale aligned image-text datasets, text-to-image has made great progress such as Glide~\citep{nichol2021glide}, DALL-E 2~\citep{ramesh2022hierarchical}, Imagen~\citep{saharia2022photorealistic} and StableDiffusion~\citep{rombach2022high}. These pretrained diffusion models have been used as a diffusion prior to promote the development of many other tasks like image editing and 3D generation.

\subsection{3D Generation Using 2D Diffusion}
With the reduced dependence on 3D data, the recent development of applying pretrained 2D text-to-image diffusion models for 3D generation has significantly progressed after the pioneer works DreamFusion~\cite{poole2022dreamfusion} and SJC~\cite{wang2023score}. The key technique is the Score Distillation Sampling (SDS) method proposed in DreamFusion which enables to use 2D diffusion models with score functions to optimize a 3D representation. Subsequently, numerous works~\cite{lin2023magic3d, metzer2023latent, chen2023fantasia3d, wang2023prolificdreamer, zhao2023efficientdreamer, zhu2023hifa} have improved text input generation results, while another research line~\cite{xu2023neurallift, melas2023realfusion, deng2023nerdi, tang2023make, qian2023magic123} focuses on 3D object reconstruction with a reference image. Compared to text-to-3D methods, image-to-3D requires generation results to closely resemble reference images. RealFusion~\cite{melas2023realfusion} and NeRDi~\cite{deng2023nerdi} extract text embedding from input images to provide additional visual cues to diffusion models, while Make-It-3D~\cite{tang2023make} uses an image captioning model for detailed text descriptions. However, both methods lack lower-level image features for detailed textures. The concurrent work, TeCH~\cite{huang2023tech}, uses both detailed text descriptions and text embedding for SDS-based 3D human generation from a single image but struggles with texture inconsistency. In contrast, we propose Ref-SDS to incorporate region-aware image guidance into the attention network of diffusion models, enabling more precise control over detailed textures.

%% file: sec/3_method.tex
\section{Method}
\label{sec:method}

\begin{figure*}[t]
  \centering
   \includegraphics[width=0.99\linewidth]{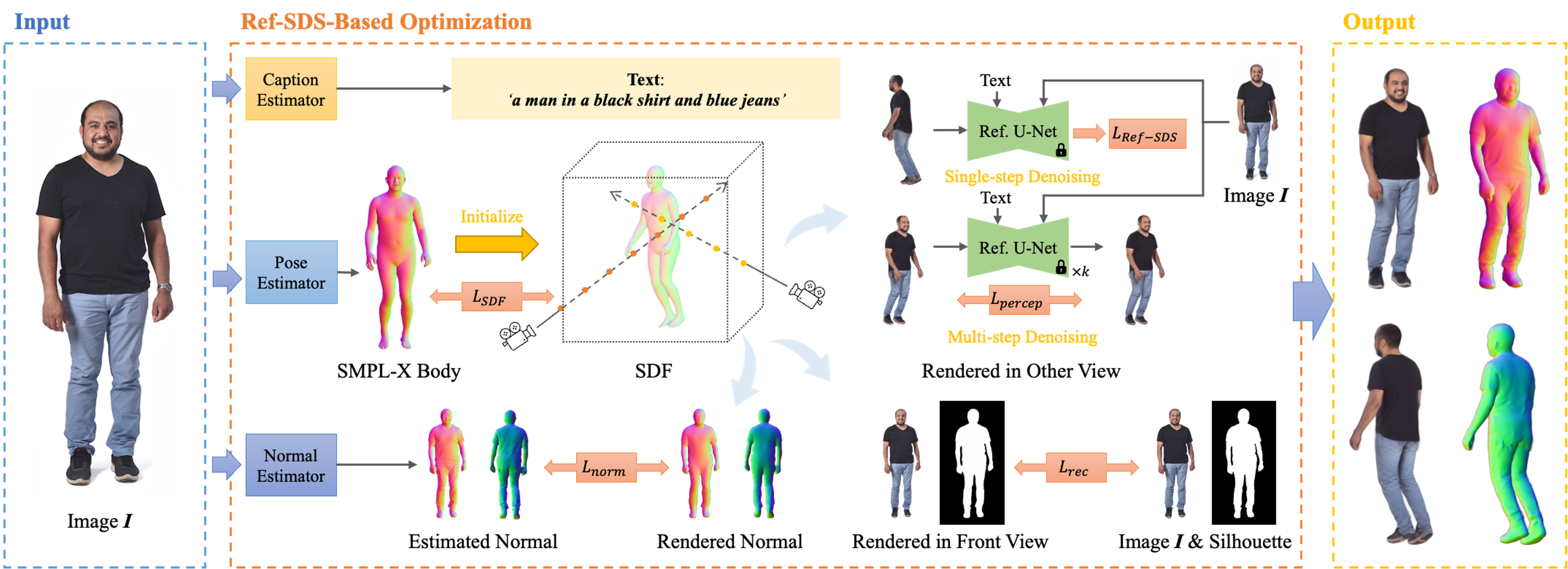}
   \caption{Overview of our proposed HumanRef for 3D clothed human generation from a single input image. }
   % \vspace{-0.1in}
   \label{fig:pipeline}
\end{figure*}

We introduce HumanRef, a unified coarse-to-fine optimization framework for 3D clothed human generation from a single image, as shown in Fig.~\ref{fig:pipeline}. 
Given an input image, we initially extract its text caption, SMPL-X body~\cite{pavlakos2019expressive}, front and back normal maps, and silhouette using estimators. A neural SDF network, initialized with the estimated SMPL-X body, is then employed for optimization-based generation. To maintain appearance and pose consistency, we use the input image, silhouette, and normal maps as optimization constraints. For invisible regions, we introduce Ref-SDS, a method that distills realistic textures from a pretrained diffusion model, yielding sharp, realistic 3D clothed humans that align with the input image.
% Given an input image, we utilize existing estimators to derive the corresponding text caption, skinned SMPL-X body, front and back normal maps, and silhouette. Subsequently, we initialize a neural SDF network with the estimated SMPL-X body and initiate an optimization-based generation process. To ensure consistency in appearance and pose between the generated 3D human and input image, we employ the input image, its silhouette, and the front and back normal maps as constraints during optimization under the reference view. For regions invisible to the input, we propose a Ref-SDS method to distill realistic textures consistent with the input image from a pretrained diffusion model. To further enhance the texture details, we apply a multi-step denoising strategy to the novel views with the diffusion model. The denoised results are then utilized to further refine the generation model. By employing these strategies, our HumanRef could generate a sharp and realistic 3D clothed human aligned with the input image.

\subsection{Image Preprocessing}
\label{sec: pre}
Creating a 3D human from a single image is inherently ill-posed, requiring effective regularization to constrain the outcomes. To tackle this challenge, we extract multiple annotations for optimization, serving as valuable constraints to enhance the accuracy of the results.

\noindent\textbf{Image Segmentation.} Given an input image $I$, we first employ Mask-RCNN~\cite{he2017mask} for background segmentation, deducing the human silhouette $S_I$. Then, we use a human parsing method~\cite{li2020self} to infer attribute masks for regions such as the head, coat, pants, etc. This enables us to divide the target human into four regions: head, upper body, lower body, and feet, each represented by their respective masks $\{M_j\}$. These regional masks will be used to instruct subsequent region-aware reference-guided generation.
% To identify the target human area in the input image, we use the Mask-RCNN~\cite{he2017mask} to segment the background from the input image $I$ and produce the human silhouette $S$. Then, we adopt a human parsing method~\cite{li2020self} to infer the mask of human attributes such as head, coat, pants, etc. Based on this, the target human is roughly divided into four regions: head, upper body, lower body, and feet, each represented by corresponding masks $\{M_j\}$. These regional masks will be used to instruct subsequent region-aware reference-guided generation.

\noindent\textbf{Image Captioning.} Similar to previous SDS-based 3D generation methods, our method is also based on a pretrained text-to-image diffusion model where text input is required. Thus, we adopt an off-the-shelf image captioning method~\cite{li2023blip} to estimate a rough text description.
% Similar to previous SDS-based 3D generation methods, we also implement distillation based on a pretrained text-to-image diffusion model where text input is required. Since our Ref-SDS method introduces additional image-level guidance, it does not heavily rely on an elaborate text description. Thus, we adopt an off-the-shelf image captioning method~\cite{li2023blip} to estimate a rough text description.

\noindent\textbf{SMPL-X Fitting.} To capture the human pose and coarse shape, we employ a human pose estimation algorithm~\cite{pavlakos2019expressive} to infer the corresponding SMPL-X body mesh from the image. We then convert this body mesh into an SDF representation and use it to initialize our SDF Network.

\noindent\textbf{Normal Estimation.} We employ the normal estimator of ECON~\cite{xiu2023econ} to predict the front and back normal maps from the image $I$. Such normal maps will provide effective geometric priors for our optimization.

\subsection{Hash-Encoded SDF Representation}
We adopt the hash-encoded SDF network~\cite{wang2023neus2} as the 3D representation of the human, with one sub-network $f_{s}$ to predict the signed distance value $s$ of a spatial query point $\mathbf{x}$ and another sub-network $f_{c}$ to predict its color $\mathbf{c}$:
\begin{equation}
s(\mathbf{x}) = f_{s}\left(\mathbf{x}, h\left(\mathbf{x}\right)\right),~~~\mathbf{c}(\mathbf{x}) = f_{c}\left(\mathbf{x}, h\left(\mathbf{x}\right)\right),
\label{equ:sdf_pred}
\end{equation}
% \begin{equation}
% s(\mathbf{x}) = f_{s}\left(\mathbf{x}, h\left(\mathbf{x}\right)\right),
% \label{equ:sdf_pred}
% \end{equation}
% \begin{equation}
% \mathbf{c}(\mathbf{x}) = f_{c}\left(\mathbf{x}, h\left(\mathbf{x}\right)\right),
% \label{equ:color_pred}
% \end{equation}
where $h\left(\mathbf{x}\right)$ is the feature queried from the hash grids with multi-level resolutions. 
% Generally, the network $f_{s}$ and $f_{c}$ share the hash features in the same levels. However, in our experiments, we find that the high-frequency information in the high-resolution hash grid helps to obtain detailed textures, but it also leads to undesired artifacts on the surface of the generated geometry. Therefore, in practice, we restrict the SDF network $f_{s}$ to query the low-resolution hash features as input, and the high-resolution hash features are only used in the color prediction network $f_{c}$.
Furthermore, we employ a cumulative distribution function~\cite{yariv2021volume} to model the density $\sigma$ from the predicted $s$:
\begin{equation}
\sigma(\mathbf{x}) = \frac{\alpha}{2}\left( 1+\textit{sign}(s(\mathbf{x}))\exp\left(-\alpha\left|s(\mathbf{x})\right|\right)\right),
\label{equ:density_pred}
\end{equation}
where $\alpha$ is a learnable parameter. $\textit{sign}(s)$ indicates the sign of the distance value $s$. Then, we can calculate the rendered pixel color $\mathbf{C}$, normal $\mathbf{N}$, and silhouette $S$ via the volume rendering method:
\begin{equation}
\Psi(\mathbf{r}) = \int_{t_n}^{t_f}T(t) \sigma(\mathbf{r}(t)) \psi(\mathbf{r}(t)) \dif t,
\label{eq:volume_render}
\end{equation}
% \begin{equation}
% \mathbf{C}(\mathbf{r}) = \int_{t_n}^{t_f}T(t) \sigma(\mathbf{r}(t)) \mathbf{c}(\mathbf{r}(t)) \dif t,
% \label{eq:volume_render_c}
% \end{equation}
% \begin{equation}
% \mathbf{N}(\mathbf{r}) = \int_{t_n}^{t_f}T(t) \sigma(\mathbf{r}(t)) \mathbf{n}(\mathbf{r}(t)) \dif t,
% \label{eq:volume_render_n}
% \end{equation}
where $\Psi$ indicates one of $\mathbf{C}$, $\mathbf{N}$, and $S$. $\psi(\mathbf{r}(t))$ is $\mathbf{c}(\mathbf{r}(t))$, $\mathbf{n}(\mathbf{r}(t))$, and $1$ for $\mathbf{C}$, $\mathbf{N}$, and $S$, respectively. $\mathbf{n}(\mathbf{x})=\nabla_{\mathbf{x}}s(\mathbf{x})$ indicates the predicted normal of query point $\mathbf{x}$. $\mathbf{r}(t) = \mathbf{o} + t\mathbf{d}$ represents the coordinate of the sampled point on the camera ray emitted from the pixel center $\mathbf{o}$ with the direction $\mathbf{d}$. $t_n$ and $t_f$ are near and far bounds of the ray, respectively. $T(t)=\exp \left( -\int_{t_n}^{t}\sigma \left(\textbf{r}(\tau)\right)\, \dif\tau\right)$ is the accumulated transmittance along the ray.
Note that, unlike previous volume rendering methods, we adopt orthographic rendering instead of perspective rendering. That is why $\mathbf{o}$ indicates the pixel center instead of the camera center.

\subsection{Reference-Guided Optimization}
In our unified reference-guided generation framework, we employ a coarse-to-fine optimization strategy. Initially, we optimize at a low rendering resolution ($64\times64$) for quick convergence to the target space. Subsequently, we progressively increase the rendering resolution up to $512\times512$, enabling the model to refine geometry and texture. Unlike previous SDS- or VSD-based methods~\cite{poole2022dreamfusion, raj2023dreambooth3d, wang2023prolificdreamer}, we propose a modified Ref-SDS method to introduce reference guidance for the 3D generation (Sec.~\ref{sec:refsds}), and region-aware attention for precise local-region guidance in Ref-SDS (\ref{sec:aware_ref}). 
% To enhance the texture, we further incorporate a multi-step denoising strategy at the later stage of optimization (Sec.~\ref{sec:multidenoise}). 
Besides, we also employ some constraints to align our generated 3D human with the input $I$ (Sec.~\ref{sec:loss}).

\begin{figure}[t]
  \centering
   \includegraphics[width=0.99\linewidth]{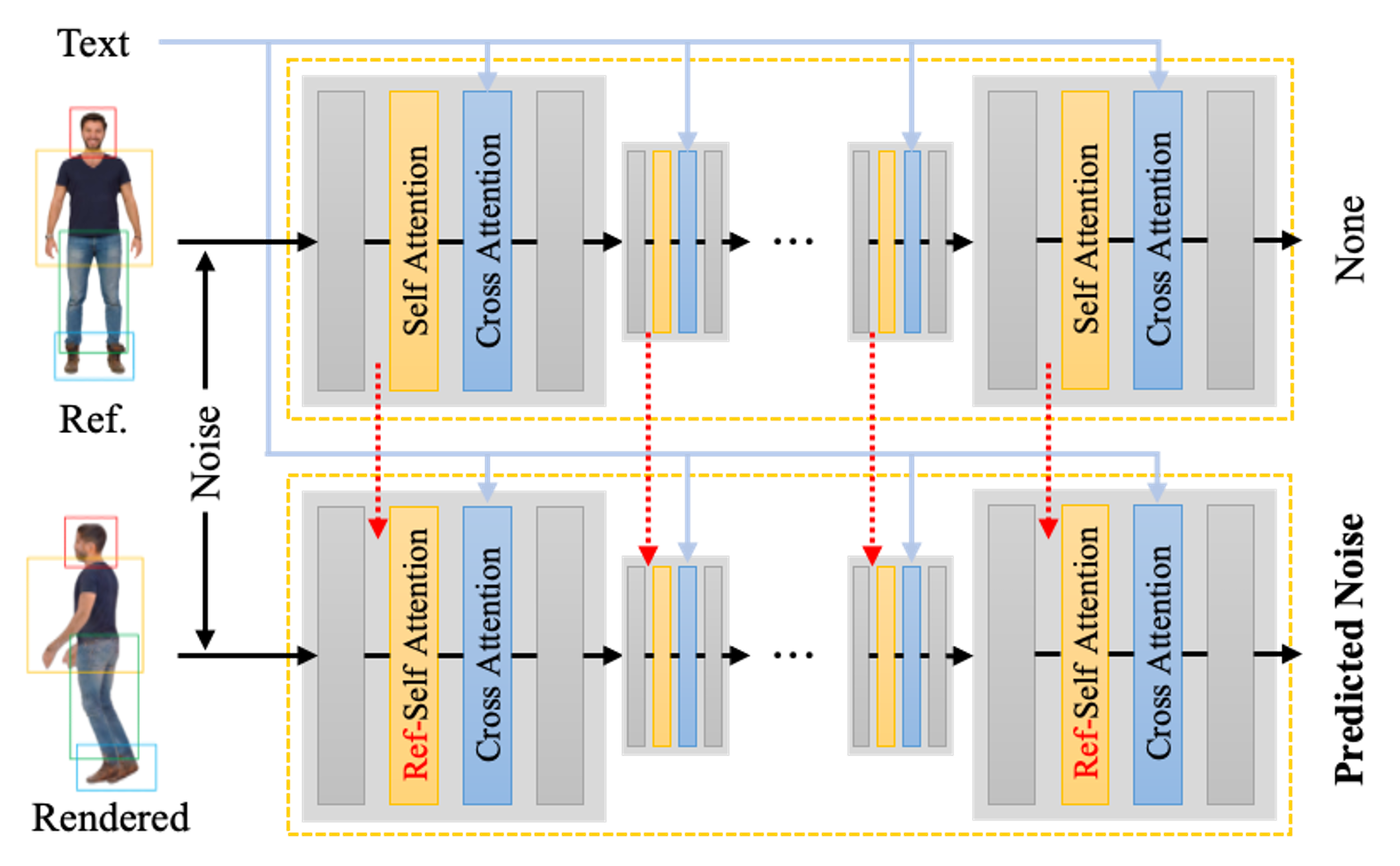}
   \caption{The framework of our reference U-Net.}
   \label{fig:ref-unet}
\end{figure}

\subsubsection{Reference-Guided Score Distillation Sampling}
\label{sec:refsds}
Inspired by~\cite{cao2023masactrl}, we modify the vanilla SDS by injecting region-aware image guidance into the attention network of U-Net during the single-step denoising process. In the implementation of vanilla SDS, a noisy rendered image, as well as random time step $t$ and text prompt, are fed into the diffusion U-Net to predict corresponding noise. Then, the weighted noise residual between predicted and added noises is regarded as the SDS loss that backpropagates to the 3D representation through the rendered image. In fact, as shown in the upper branch of Fig.~\ref{fig:ref-unet}, the noisy image will undergo several transformer blocks and up/down-sampling processes in the denoising U-Net. In each transformer block, the latent feature derived from the input is first subjected to self-attention operation and then cross-attention operation with text and time embedding, thereby realizing the text-guided denoising process. Note that we do not draw the data flow of time embedding in the figure because we do not make any changes to this part. To achieve the reference image guidance, as shown in Fig.~\ref{fig:ref-unet}, we first perform the diffusion and denoising process normally on the reference image and save the latent features $p^{ref}_i$ before each self-attention operation. Subsequently, we feed the noisy rendered image to the same U-Net to perform denoising. Instead of undergoing the self-attention and cross-attention, we concatenate the latent features $p^{tar}_i$ before self-attention with the corresponding features $p^{ref}_i$ derived from the denoising process of the reference image, and perform ref-self-attention between the original features $p^{tar}_i$ and the concatenated features $p^{tar}_i\cup p^{ref}_i$. In this way, the information from the reference image is transferred to the denoising process of the rendered image, thereby our denoising U-Net achieves reference image guidance.

\begin{figure}[t]
  \centering
   \includegraphics[width=0.99\linewidth]{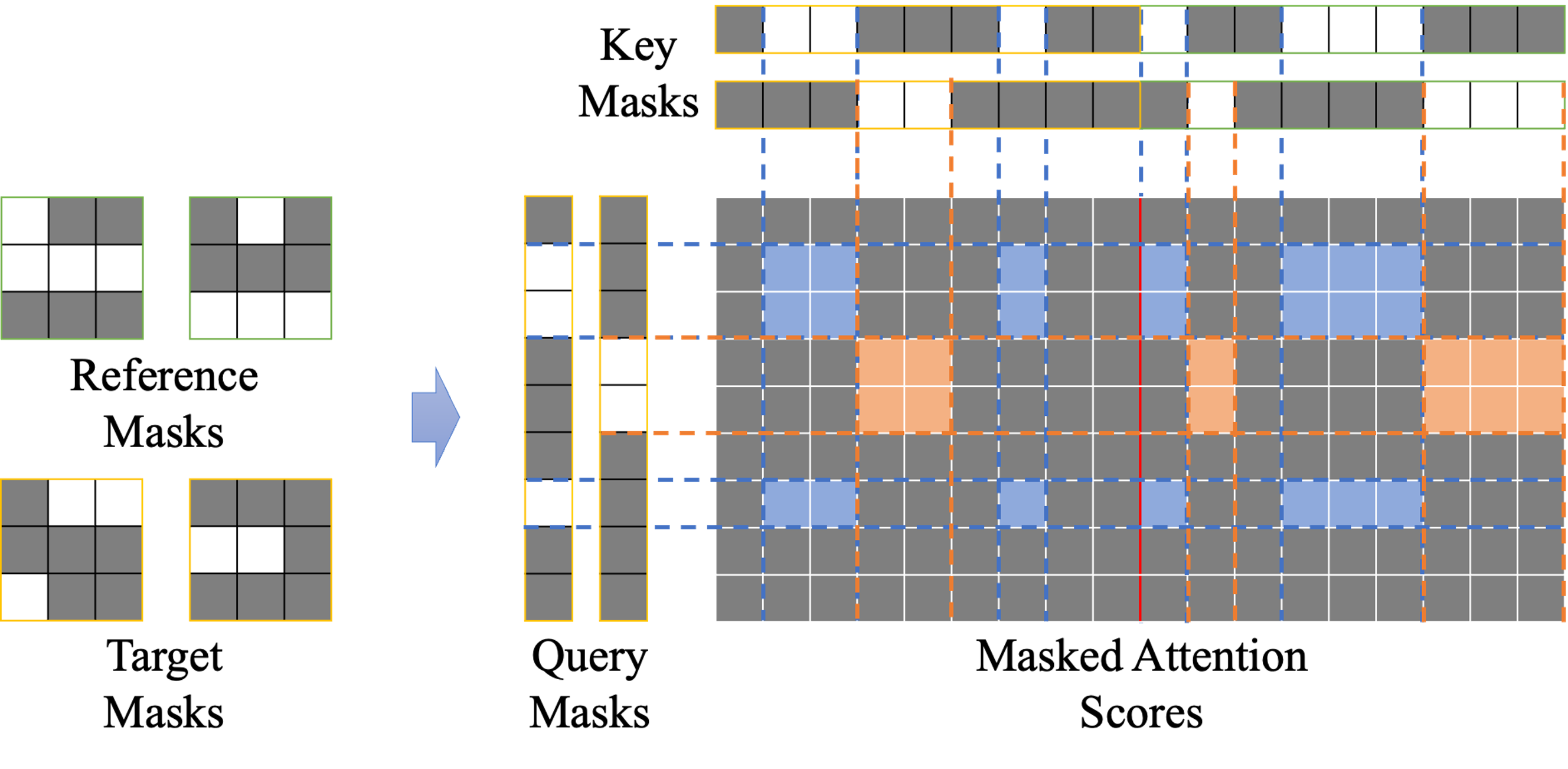}
   \caption{Example of Region-Aware Attention.}
   \label{fig:region_atten}
\end{figure}

\subsubsection{Region-Aware Attention}
\label{sec:aware_ref}
The mutual process in Sec.~\ref{sec:refsds} only provides global attention to the denoising process of the rendered image. To achieve region-aware attention, we further introduce additional attention masks into the above process. In fact, we have inferred the regional masks $\{M_j\}$ of reference image in preprocessing. To obtain the regional masks $\{\Bar{M}_j\}$ of the rendered image, we first divide the 3D space into different regions in the vertical direction according to the boundaries of region masks $\{M_j\}$, and then divide the rendered image into corresponding regions according to the projection of the space division in the rendering view. Subsequently, we feed these regional masks into the denoising process for region-aware attention.

% \xy{Specifically, we first divide the 3D space into different regions in the vertical direction according to the region masks $\{M_j\}$ of the input image $I$ (mentioned in Sec.~\ref{sec: pre}). [Xiaoyu: unclear]} According to the division of spatial regions and the silhouette of the rendered view, we can divide the rendered image into corresponding regions, i.e., the corresponding region masks $\{\Bar{M}_j\}$ of the rendered image. 
% Then, we utilize these masks in the denoising process of both reference and rendered images to achieve region-aware self-attention and ref-self-attention. 

As shown in Fig.~\ref{fig:region_atten}, we implement the region-aware attention on the attention score matrix calculated from \textit{query} vector and \textit{key} vector. Assuming a $3\times3$ latent feature and its two attention masks, to implement ref-self-attention, the target feature and its masks are flattened into $1\times9$ vectors. After passing a linear network, the feature vector is then used as the \textit{query} vector to calculate the attention score matrix together with the \textit{key} vector formed by concatenating the target and reference features. Correspondingly, we use the flattened masks and the concatenated masks as the \textit{query} and \textit{key} masks to infer the local regions with same semantics, as shown in the blue and orange regions of Fig.~\ref{fig:region_atten}. After determining the local attention regions, we multiply attention scores outside these regions by a coefficient $\gamma<1$, and then normalize the whole score matrix, thereby improving the network's attention on these determined local regions. In practice, we set  $\gamma$ as $0.3$.

After that, we now implement the denoising process with region-aware reference image guidance. Continuing from vanilla SDS~\cite{poole2022dreamfusion}, we also use the weighted noise residual between predicted and added noises as our Ref-SDS for 3D generation. Thanks to the image-level guidance, 3D generation based on our Ref-SDS is robust, thus we could set the CFG scale in the common level as the image generation. In this way, our Ref-SDS supports the generation of realistic textures that are consistent with the input reference image.

\subsubsection{Loss Functions}
\label{sec:loss}
During the optimization, we render our SDF network in front, back, and other random views. In the front view, we construct a reconstruction loss formed as $L_{rec}=L_{rgb}+L_{IoU}$. Here, 
$L_{rgb}$ is a $L_2$-form loss calculated between the rendered and input color images. $L_{IoU}$ indicates the intersection-over-union (IoU) loss \cite{zhang2022adaptive} between the rendered silhouette $S$ and the silhouette $S_I$ of input image:
\begin{equation}
L_{IoU}=1-\frac{\left\|S \otimes S_I\right\|_{1}}{\left\|S \oplus S_I - S \otimes S_I \right\|_{1}},
\label{eq:iou}
\end{equation}
where $\otimes$ and $\oplus$ indicate element-wise product and sum operator, respectively. Furthermore, we calculate a $L_2$-form normal loss $L_{norm}$ between the rendered and estimated normal maps in the front and back views. For views other than the front view, we implement the proposed Ref-SDS loss $L_{Ref-SDS}$ to generate realistic texture and geometry. To further enhance the texture details, we implement a multi-step denoising process based on our region-aware reference-guided U-Net, and calculate the perceptual loss~\cite{johnson2016perceptual} $L_{percep}$ between the rendered image and the enhanced image.
% and the perceptual loss~\cite{johnson2016perceptual} $L_{percep}$.
To constrain the pose of the generated human, we additionally introduce a $L_1$-form SDF loss $L_{SDF}$ between the predicted signed distance value $s$ and that queried from SMPL-X body mesh. Besides, we also adopt the normal smooth loss $L_{smooth}$ as the previous generation methods \cite{tang2023make, liu2023zero} during the optimization process.

In summary, we train the entire network using the following objective function:
\begin{equation}
\begin{split}
L= & \lambda_{1} L_{rec} + \lambda_{2} L_{norm} + \lambda_{3} L_{Ref-SDS}\\
+ & \lambda_{4} L_{percep} + \lambda_{5} L_{SDF} + \lambda_{6} L_{smooth},
\end{split}
\end{equation}
where $\{\lambda_{1}, \dots ,\lambda_{6}\}$ are the weights used to balance different loss terms. In practice, we empirically set the weights as $10000, 100, 0.001, 20, 100, 5$, respectively.
% We have omitted the weights used to balance different loss terms for simplification and give detailed information regarding weight settings in Sec.~\ref{sec:details}.

\subsection{Implementation Details}
\label{sec:details}
% 随机背景颜色、随机渲染视角的选择、stable diffusion版本\A100\
We implement our HumanRef with the ThreeStudio~\cite{threestudio2023} framework in Pytorch~\cite{paszke2019pytorch} on a A100 GPU. For optimization, we adopt the Adam~\cite{kingma2014adam} optimizer with default hyperparameters and a learning rate of $0.001$ for all learnable parameters.
% including the features in the hash grid, SDF network $f_s$, color network $f_c$, and the parameter $alpha$ in Eq.~\ref{equ:density_pred}. 
To guarantee the quality of the generated model, we uniformly sample random views in the elevation range $[-20, 20]$ and azimuth range $[-180, 180]$ outside the front and back views. In addition, to facilitate the model to distinguish generated foreground objects and background, we follow \cite{tang2023make} to set a random background color for the rendering results in each optimization step. To perform region-aware Ref-SDS generation and multi-step denoising, we adopt the stable diffusion model in version 1.5~\cite{rombach2022high} and the fast diffusion sampling scheduler UniPC~\cite{zhao2023unipc}. 
% Nonetheless, our method is also applicable to other higher version stable diffusion models.

%% file: sec/4_exp.tex
\section{Experiments}
\label{sec:exp}
% In this section, we first briefly introduce the experiment setting including baseline methods, evaluation metrics, and data source in Sec.~\ref{sec:set}. Then, we implement our HumanRef on different input images and qualitatively and quantitatively compare our results with those produced by state-of-the-art baseline methods in Sec.~\ref{sec:compari}. Furthermore, we perform ablation studies to demonstrate the effectiveness of the main components of our method.

\begin{figure*}[t]
  \centering
   \includegraphics[width=0.99\linewidth]{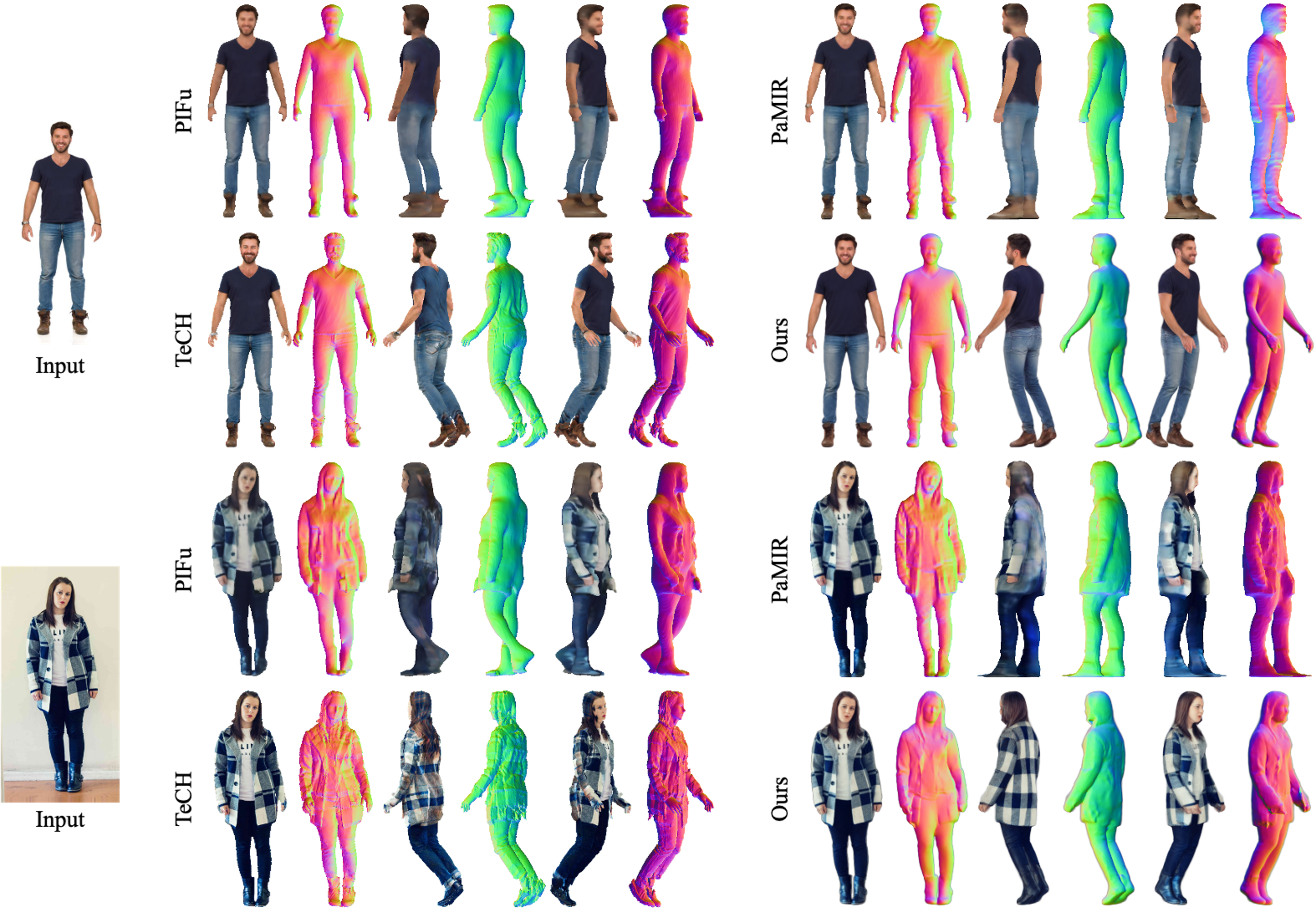}
   \caption{Qualitative comparison of results produced by baseline methods and ours.}
   \label{fig:compari}
\end{figure*}

\subsection{Setup}
\label{sec:set}
\noindent\textbf{Baseline Methods}
% PIFu, PaMIR, Make-it-3D, TeCH
To evaluate the performance of our HumanRef on 3D clothed human generation, we compare our method with three state-of-the-art baseline methods, including PIFu~\cite{saito2019pifu}, PaMIR~\cite{zheng2021pamir}, and a concurrent work TeCH~\cite{huang2023tech}. Here, PIFu and PaMIR are two learning-based methods that acquire the ability to infer the geometry and texture of a 3D clothed human from input images after being training on a large number of scanned human datasets. TeCH is a multi-stage optimization algorithm based on the SDS method designed for 3D clothed human generation. For evaluation purposes, we utilize human images released by \cite{fu2022stylegan, alldieck2022photorealistic, xiu2023econ}.

\noindent\textbf{Evaluation Metrics}
Following~\cite{tang2023make}, we adopt LPIPS~\cite{zhang2018unreasonable} between the input image and rendered image at the input view to evaluate the reconstruction quality. Besides, we use the contextual distance~\cite{mechrez2018contextual} and CLIP score~\cite{radford2021learning} as the generation quality metrics to measure the texture similarity and semantic similarity between the input image and rendered images at novel views.

\begin{table*}[t]
  \centering
  \begin{tabular}{@{}lcccccccc@{}}
    \toprule
    Methods & PIFu & PaMIR & TeCH & \thead{\textbf{Ours} \\ (Full)} & \thead{Ours \\ w/o $L_{rec}$} & \thead{Ours \\ w/o $L_{SDF}$} & \thead{Ours \\ w/o $L_{norm}$} & \thead{Ours \\ w/o $L_{smooth}$}\\
    \midrule
    LPIPS $\downarrow$ & 0.054 & 0.050 & 0.044 & \textbf{0.032} & 0.104 & 0.034 & 0.034 & 0.033\\
    Contextual $\downarrow$ & 3.180 & 2.961 & 2.882 & \textbf{1.969} & 4.181 & 2.480 & 2.447 & 2.304\\
    CLIP Score $\uparrow$ & 80.1\% & 81.4\% & 85.3\%  & \textbf{90.0}\% & 78.9\% & 88.7\%  & 88.8\% & 88.5\%\\
    \bottomrule
  \end{tabular}
  \caption{Quantitative comparison of results produced by baselines and ours. Compared with baseline methods, our HumanRef achieves lower metric scores on LPIPS and contextual distance, and higher CLIP scores on semantic similarity. This means that our method is able to generate 3D clothed human that considers both reconstruction and generation quality.}
  \label{tab:exp_all}
\end{table*}

\subsection{Comparisons}
\label{sec:compari}

We evaluate our HumanRef and baseline methods for 3D clothed human generation on diverse images with various clothing styles and textures, as shown in Fig.~\ref{fig:compari}. Additionally, we provide quantitative comparison results to assess the performance of different methods in terms of reconstruction and generation qualities, as shown in Tab.~\ref{tab:exp_all}. Clearly, our method surpasses the baseline methods in both qualitative and quantitative comparisons. 

Contrary to TeCH and ours, PIFu and PaMIR employ 3D generators trained on scanned human datasets to predict human body geometry and texture. Their performance thus is limited by the training data and model design, struggling to infer detailed textures and fine geometry from a single image, particularly in areas invisible to the input. The first and third rows in Fig.~\ref{fig:compari} illustrate the disparity between front and back views, with the latter appearing blurrier and less detailed. As a result, they receive lower evaluation scores shown in Tab.~\ref{tab:exp_all}.
TeCH, however, can generate detailed textures in unseen areas due to SDS-based text-guided optimization and pretrained diffusion model priors. To mitigate over-saturation in SDS-based generation and enhance realism, TeCH employs strategies to minimize SDS denoising diversity, including precise text inference via a question-answering algorithm, textural invention, and diffusion model fine-tuning. Nonetheless, TeCH struggles to achieve texture consistency with the input image due to the lack of image-level guidance in SDS denoising. In the \textit{Man} example of Fig.~\ref{fig:compari}, TeCH generates a realistic blue T-shirt texture on the back, but there is still a noticeable difference from the dark blue texture in the input image. In contrast, our method, HumanRef, generates realistic and detail-rich textures that are more consistent with the input image. For geometry generation, although lacking a specialized design like TeCH, our method produces reasonable and complete 3D human structures, while TeCH suffers from issues like geometric breakage and stretching in the generated human feet. Moreover, in the \textit{Woman} example of Fig.~\ref{fig:compari}, our method successfully generates realistic view-consistent texture and geometry for humans with complex clothing textures that TeCH fails to infer.

\begin{figure}[t]
  \centering
   \includegraphics[width=0.99\linewidth]{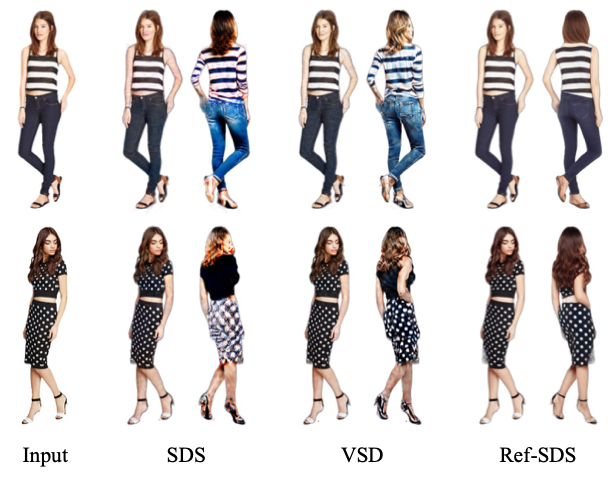}
   \caption{3D cloth human generation based on SDS, VSD, and our Ref-SDS, with estimated text prompts '\textit{a woman in a striped top and jeans}' and '\textit{a woman in black and white polka dot print skirt}'.}
   \label{fig:vs_sds}
\end{figure}

\begin{figure}[t]
  \centering
   \includegraphics[width=0.99\linewidth]{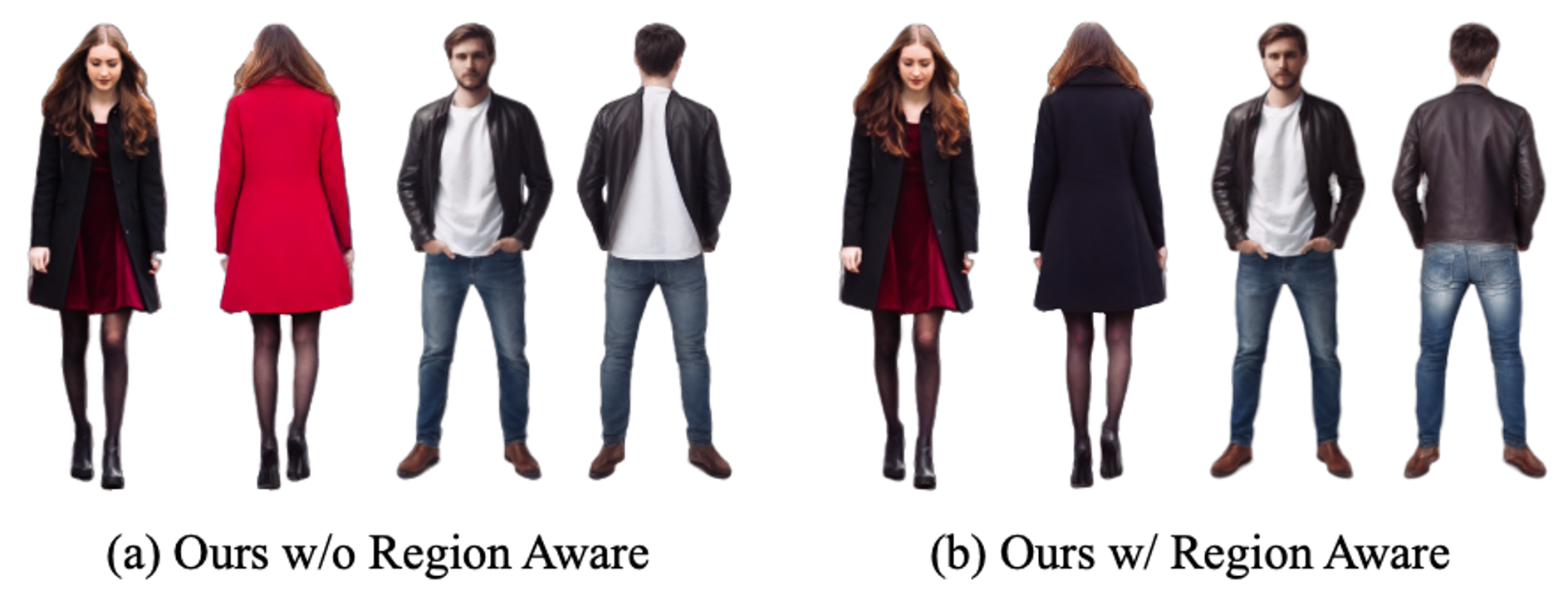}
   \caption{Ablation study on the region-aware attention.}
   \label{fig:aware}
\end{figure}

\subsection{Ablation Studies}
\label{sec:abla}
\noindent\textbf{Ref-SDS.}
To illustrate the superiority of the proposed Ref-SDS in our framework, we conduct a comparative experiment by replacing it with exiting SDS~\cite{poole2022dreamfusion} and VSD~\cite{wang2023prolificdreamer} methods. Here, SDS and VSD are both text-guided methods, with VSD being a modified version of SDS to address over-saturation. In Fig.~\ref{fig:vs_sds}, SDS generates textures with higher saturation and lacks realism, while VSD produces realistic textures that are not consistent with the input. In contrast, our Ref-SDS-based method generates realistic and view consistent textures thanks to the reference image guidance in the denoising process.

\noindent\textbf{Region-Aware Attention.}
Additionally, we evaluate the impact of the region-aware attention in our Ref-SDS by comparing our full framework with a version without it. In Fig.~\ref{fig:aware}, we observe that the implementation without region-aware attention may spontaneously focus on undesired features in the reference image during ref-self-attention, resulting in realistic but unreasonable textures in some cases. By incorporating region-aware attention, we effectively limit the attention area of the network and obtain more reasonable texture inference results.

% Here, SDS and VSD are both implemented based on text guidance while VSD is regarded as a modified version of SDS for solving its over-saturation problem. As shown in Fig.~\ref{fig:vs_sds}, although our framework based on SDS or VSD enable to generate 3D clothed human from a single image, they cannot generate real textures that are consistent with the input image due to the lack of reference image guidance in the denoising process. In comparison, the texture generated based on SDS has higher saturation and lacks realism, while the result generated based on VSD is realistic but not consistent with the input. By contrast, our Ref-SDS-based method supports to generate realistic textures consistent with the inputs. 

% Furthermore, to validate the effect of the region-aware attention mechanism in our Ref-SDS, we compare our full framework with the one without region-aware attention. As shown in Fig.~\ref{fig:aware}, we find that the implementation without region-aware attention may spontaneously focus on some undesired local features in the reference image during the ref-self-attention, thereby predicting realistic but unreasonable textures in some cases. By introducing region-aware attention, we can effectively limit the attention area of the network and obtain more reasonable texture inference results.

\begin{figure}[t]
  \centering
   \includegraphics[width=0.99\linewidth]{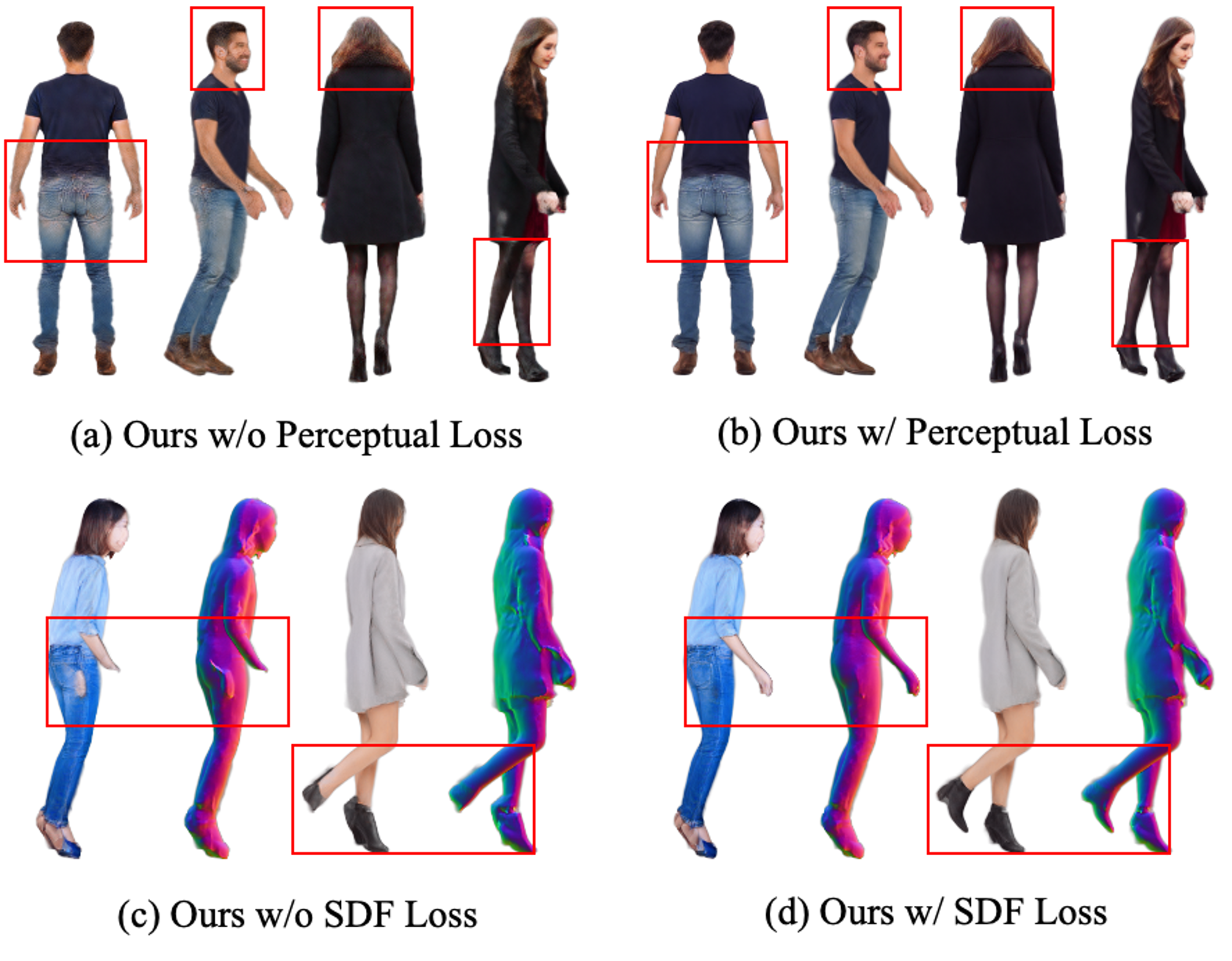}
   \caption{Ablation study on analyzing perceptual loss for texture enhancement and SDF loss for body pose preservation.}
   \label{fig:multi_sdf}
\end{figure}

\noindent\textbf{Loss Functions.}
To assess the effectiveness of our core loss functions, we perform experiments by systematically removing them and evaluating the corresponding results. In the red box of Fig.~\ref{fig:multi_sdf}(a), the method without the perceptual loss produces blurred texture details, although it remains realistic and consistent with the input. By incorporating the perceptual loss, we significantly enhance the local details of the generated human, as depicted in Fig.~\ref{fig:multi_sdf}(b). Furthermore, Fig.~\ref{fig:multi_sdf}(c\&d) and Tab.~\ref{tab:exp_all} demonstrate the constraining role of the SDF loss in preserving human geometric pose and the necessity of retaining other loss designs, including reconstruction loss, normal loss, and smoothing loss, to improve the robustness and generation quality of our method.

% To investigate the effectiveness of our core loss functions, we further perform experiments by removing them one by one and evaluate the corresponding results. As shown in the red box of Fig.~\ref{fig:multi_sdf}(a), the method without the multi-step denoising strategy results in blurred texture details although it is realistic and consistent with the input. Through the constraint of perceptual loss used in the multi-step denoising strategy, the local details of the generated human are significantly enhanced, as shown in Fig.~\ref{fig:multi_sdf}(b). Moreover, as shown in Fig.~\ref{fig:multi_sdf}(c\&d) and Tab.~\ref{tab:exp_all}, we illustrate the constraining role of the SDF loss in preserving human geometric pose, and the necessity of retaining other loss designs to improve the robustness and generation quality of the method, including reconstruction loss, normal loss, and smoothing loss.

%% file: sec/5_conclusion.tex
\section{Conclusion}
\label{sec:conclusion}

\begin{figure}[t]
  \centering
   \includegraphics[width=0.99\linewidth]{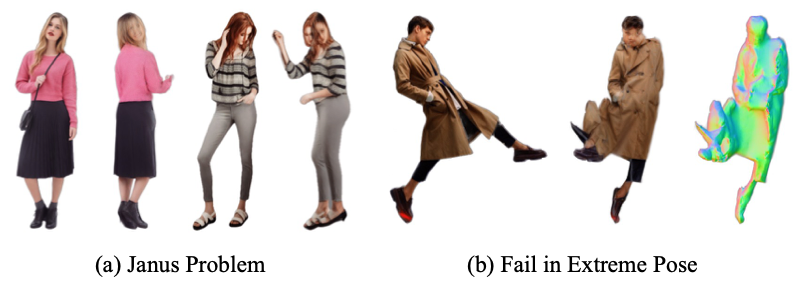}
   \caption{Our approach may suffer from the Janus problem in the side view because we do not make any constraints on this in the side view. Besides, our method may also fail in some extreme poses where SMPL-X body estimation is wrong.}
   \label{fig:limi}
\end{figure}

In this paper, we propose HumanRef, a unified framework for generating 3D clothed humans from a single input image. Our approach addresses the challenge of realistic texture generation in the invisible areas by proposing a modified Ref-SDS method that fully exploits the guidance of the reference image during the denoising process. Additionally, we introduce region-aware attention into our Ref-SDS, enhancing the precision of image guidance. Overall, our HumanRef framework empowers the generation of 3D clothed humans with view-consistent realistic textures and reasonable geometry from a single reference image.

% By injecting the reference image into the diffusion model, we generate view-consistent results that preserve the visual appearances of the reference image.
% inject the image-level guidance into the denoising-based 3D generation. Moreover, we introduce a multi-step denoising strategy to enhance the generated texture details. With these advancements, our HumanRF generates a 3D clothed human with view-consistent realistic textures and reasonable geometry from a single input image.

\noindent\textbf{Limitation.} 
While our 3D human generation experiments have achieved impressive overall results, it is crucial to acknowledge occasional failures. Fig.~\ref{fig:limi}(a) visually exemplifies the Janus problem, which can impact our results in side views due to the lack of view-specific constraints. Moreover, accurately estimating body poses in extreme cases challenges for our method, potentially resulting in failures, shown in Fig.~\ref{fig:limi}(b).
% Although our 3D human generation experiments have yielded impressive results overall, it is important to acknowledge the existence of some instances where failures occurred. As shown in Fig.~\ref{fig:limi}, the Janus problem may affect our approach in the side view due to the absence of constraints specifically designed for this perspective. Additionally, our method may encounter difficulties in accurately estimating body poses in extreme cases, leading to potential failures, particularly when dealing with SMPL-X body estimation.